\documentclass[acmtog,screen,authorversion]{acmart}
\acmSubmissionID{335}

\usepackage{booktabs}
\usepackage{multirow}
\usepackage{times}
\usepackage{epsfig}
\usepackage{graphicx}
\usepackage{amsmath}
\usepackage{amssymb}
\usepackage{overpic}
\usepackage{enumitem}
\usepackage{layouts}
\usepackage{hyphenat,balance,microtype}
\usepackage[fleqn,tbtags]{mathtools}
\usepackage[capitalise]{cleveref}
\usepackage[detect-weight]{siunitx}
\usepackage{caption, subcaption, textpos}
\usepackage{colortbl}
\usepackage{algorithm}
\usepackage{algorithmic}
\usepackage{verbatim}
\usepackage{listings}
\usepackage{wrapfig}

\graphicspath{{./images/}}

\newcommand{\tb}[1]{\textbf{#1}}

\newcommand{\tablestyle}[2]{\setlength{\tabcolsep}{#1}
                            \renewcommand{\arraystretch}{#2}
                            \centering
                            \footnotesize}

\newcolumntype{x}[1]{>{\centering\arraybackslash}p{#1pt}}
\newcolumntype{y}[1]{>{\raggedright\arraybackslash}p{#1pt}}
\newcolumntype{z}[1]{>{\raggedleft\arraybackslash}p{#1pt}}
\definecolor{graycolor}{gray}{.9}

\acmJournal{TOG}
\acmYear{2023}
\acmVolume{42}
\acmNumber{4}
\acmArticle{1}
\acmMonth{8}
\acmPrice{15.00}
\acmDOI{10.1145/3592131}
\setcopyright{acmlicensed}

\received{January 2023}
\received[accepted]{March 2023}
\received[final version]{May 2023}

\title{OctFormer: Octree-based Transformers for 3D Point Clouds}

\author{Peng-Shuai Wang}
\affiliation{
 \institution{Peking University}
 \country{China}
}
\email{wangps@hotmail.com}

\begin{abstract}
  We propose octree-based transformers, named OctFormer, for 3D point cloud learning.
  OctFormer can not only serve as a general and effective backbone for 3D point cloud segmentation and object detection but also have linear complexity and is scalable for large-scale point clouds.
  The key challenge in applying transformers to point clouds is reducing the quadratic, thus overwhelming, computation complexity of attentions.
  To combat this issue, several works divide point clouds into non-overlapping windows and constrain attentions in each local window.
  However, the point number in each window varies greatly, impeding the efficient execution on GPU.
  Observing that attentions are robust to the shapes of local windows, we propose a novel octree attention, which leverages sorted shuffled keys of octrees to partition point clouds into local windows containing a fixed number of points while permitting shapes of windows to change freely.
  And we also introduce dilated octree attention to expand the receptive field further.
  Our octree attention can be implemented in 10 lines of code with open-sourced libraries and runs 17 times faster than other point cloud attentions when the point number exceeds $200k$.
  Built upon the octree attention, OctFormer can be easily scaled up and achieves state-of-the-art performances on a series of 3D segmentation and detection benchmarks, surpassing previous sparse-voxel-based CNNs and point cloud transformers in terms of both efficiency and effectiveness.
  Notably, on the challenging ScanNet200 dataset, OctFormer outperforms sparse-voxel-based CNNs by 7.3 in mIoU.
  \emph{Our code and trained models are available at \url{https://wang-ps.github.io/octformer}}.
\end{abstract}

\begin{teaserfigure}
  \centering
  \includegraphics[width=0.95\linewidth]{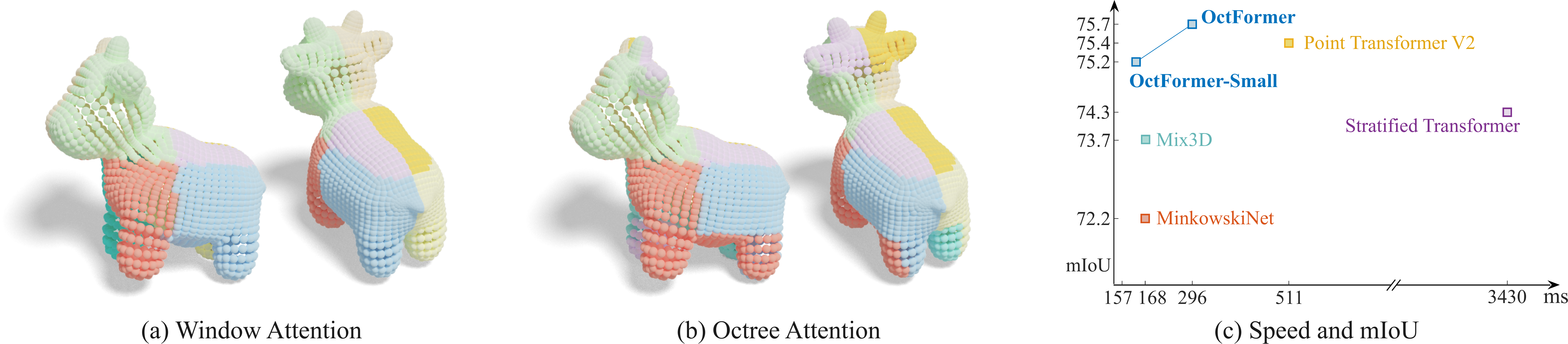}
  \caption{Octree Attention and the superiority of OctFormer.
  (a): The window attention partitions the point cloud with cubic windows and constrains the attention in each window to accelerate the global attention.
  Each window is encoded by a specific color, as indicated by the front and back view of the point cloud.
  The point number in each window is highly unbalanced, which incurs great computation cost.
  (b): Our octree attention partitions the point cloud according to the sorted shuffled keys of the octree, ensuring an equal number of points in each window.
  (c): OctFormer built upon our octree attention achieves the best mIoU and efficiency on ScanNet compared with representative sparse-voxel-based CNNs and point cloud transformers.
  The horizontal axis represents the time of one forward pass of each network on an Nvidia 3090 GPU, taking a batch of 250k points.
  }
  \label{fig:teaser}
\end{teaserfigure}

\ccsdesc[500]{Computing methodologies~Shape analysis}
\ccsdesc[500]{Computing methodologies~Point-based models}
\ccsdesc[500]{Computing methodologies~Neural networks}

\keywords{Point Clouds, Transformers, Octree, 3D Semantic Segmentation, 3D Object Detection}

\begin{document}
\maketitle

\section{Introduction} \label{sec:intro}

3D point cloud understanding is a fundamental task in computer graphics and vision and has a broad range of applications, including robotics, autonomous driving, and augmented reality.
A variety of deep learning methods have been proposed for it, such as voxel-based CNNs~\cite{Wu2015,Wang2017,Graham2018}, view-based CNNs~\cite{Su2015}, and point-based networks~\cite{Qi2016,Qi2017,Li2018}, and remarkable progress has been made.
Recently, point cloud transformers have emerged~\cite{Guo2021,Zhao2021,Misra2021} as an effective alternative with the potential for cross-multimodality training and general intelligent models~\cite{Radford2021,Ramesh2022}.

However, the efficiency of point cloud transformers is still much worse than their CNN counterparts~\cite{Wang2017,Choy2019,Nekrasov2021}, especially on scene-scale datasets like ScanNet~\cite{Dai2017a}, and the performance of point cloud transformers is also just comparable.
Since it has been proven that transformers are at least as expressive as CNNs~\cite{Cordonnier2020},
one of the key challenges of applying transformers to point clouds is to overcome the huge computational complexity of transformers, which is quadratic with the number of elements involved.
Several methods~\cite{Guo2021,Yu2022,Pang2022} directly apply transformers to all points globally, thus limiting their applicability to large-scale point clouds.
Following the progress in scaling up vision transformers~\cite{Liu2021a,Dong2022}, one effective strategy is to constrain point cloud transformers within non-overlapping local windows~\cite{Lai2022,Sun2022,Fan2022,Mao2021}.
However, unlike images, the number of points across different local windows varies significantly due to the sparsity of point clouds.
To deal with this issue, sophisticated implementations like region batching~\cite{Fan2022,Sun2022} or customized GPU kernels~\cite{Lai2022} have to be adopted, which severely impedes massive parallelism on GPUs.
Another strategy to speed up point cloud transformers is to apply transformers in downsampled feature maps~\cite{Park2022,Cheng2022}, which also weakens the network capability and incurs a decrease in performance.

In this paper, we present a general and scalable octree-based transformer, abbreviated as OctFormer, for learning on 3D point clouds.
The key building block of OctFormer is a novel octree attention mechanism for point clouds.
To retain linear complexity, we divide each point cloud into small groups when applying attentions.
Our key observation is that attentions are insensitive to the actual shape of underlying local windows.
Instead of using cubic windows as in previous works, which incur variant point numbers in each window, we divide point clouds into groups with irregular windows while keeping the point number in each window the same.
Consequently, we can easily implement our attention using standard operators provided by deep learning frameworks like PyTorch~\cite{Paszke2019}.
To generate the required window partition, our second observation is that after constructing an octree with the parallel algorithm in~\cite{Zhou2011}, the octree nodes are sorted in z-order by shuffled keys~\cite{Wilhelms1992}, which ensures that spatially-close octree nodes are contiguously stored in memory.
We store features in tensors according to the order of octree nodes.
After padding a few zeros to make the spatial numbers of tensors divisible by the specified point number in each window, we can efficiently generate the window partition by simply reshaping the tensors at almost zero cost.
An example is shown in Figure~\ref{fig:teaser}-(b), where the point number in each window is the same.
To further increase the receptive fields of OctFormer, we introduce a dilated octree attention with dilated partitions along the spatial dimension of tensors, which can also be efficiently implemented with tensor reshaping and transposing.

Our OctFormer challenges conventional wisdom in designing point cloud transformers from two aspects. 
First, instead of using fixed-sized local windows, we fix the point number in each window when doing point cloud partition, enabling simple implementation and easy parallelization; 
second, instead of regarding point clouds as unordered and unstructured point sets, we actually sort the quantized points with shuffled keys by building octrees, resulting in a convenient window partition.
Our octree attention completely eliminates the expensive neighborhood searching used in previous designs~\cite{Wu2022,Lai2022}, bypasses the sparsity of point clouds, and can be reduced to a standard multi-head self-attention~\cite{Vaswani2017} on small groups of equal size. 
Consequently, our octree attention can be implemented in \emph{10 lines of code} with open-sourced libraries freely available on the web. 
One \emph{single} transformer block on top of octree attention runs at least \emph{17 times faster} than previous state-of-the-art point transformer blocks~\cite{Lai2022,Wu2022} when the number of elements involved is $200k$.

We also introduce feature hierarchies following the multiscale structure of octrees, endowing OctFormer with the capability as a general backbone for 3D segmentation and detection.
We verify the effectiveness of OctFormer on a series of 3D benchmarks.
Specifically, our OctFormer achieves the best performance on the validation set of ScanNet segmentation~\cite{Dai2017a}, SUN RGB-D detection~\cite{Song2015}, and ScanNet200 segmentation~\cite{Rozen2022}, surpassing all previous state-of-the-art sparse-voxel-based CNNs~\cite{Choy2019,Wang2017,Graham2018} and point cloud transformers~\cite{Lai2022,Wu2022} by a large margin.
Notably, on ScanNet200 segmentation, which contains 200 semantic categories (ten times more than ScanNet), the mIoU of our OctFormer is higher than MinkowskiNet~\cite{Choy2019} by \emph{7.3} and even higher than the recently-proposed LGround~\cite{Rozen2022} by \emph{5.4}, which pretrains a sparse-voxel-based CNN with a powerful CLIP model~\cite{Radford2021}.

In summary, our main contributions are as follows:
\begin{itemize}[leftmargin=10pt,itemsep=2pt]
  \item[-] We propose a novel octree attention and its dilated variant, which are easy to implement and significantly more efficient than previous point cloud attentions;
  \item[-] We propose OctFormer, which can serve as a general backbone for 3D point cloud segmentation, detection, and classification;
  \item[-] OctFormer achieves state-of-the-art performances on a series of 3D segmentation and detection benchmarks, and the computational efficiency of OctFormer is much higher than previous point cloud transformers and even surpasses highly optimized sparse-voxel-based CNNs.
\end{itemize}

\section{Related Work} \label{sec:related}

\paragraph{Voxel-based CNNs}
Full-voxel-based CNNs generalize 2D CNNs to 3D learning by representing 3D data with uniformly-sampled voxels~\cite{Wu2015,Maturana2015,Qi2016}.
However, these methods can only take low-resolution voxels like $32^3$ as input due to their cubic computational and memory cost with regard to the voxel resolution.
Sparse-voxel-based CNNs greatly improve the efficiency of full-voxel-based CNNs by constraining CNN operations into non-empty sparse voxels and adopting octrees~\cite{Wang2017,Wang2018a} or hash tables~\cite{Graham2018,Shao2018,Choy2019} to facilitate neighborhood searching for convolutions.
Several other networks also leverage octrees~\cite{Riegler2017,Riegler2017a,Lei2019} for improving the efficiency of full-voxel-based CNNs.
Sparse-voxel-based CNNs mainly use small convolution kernels, while it has been proven that large convolution kernels can greatly improve network performance~\cite{Chen2022} in practice.
Our OctFormer can easily enlarge its window size and has a much larger receptive field than sparse-voxel-based CNNs.

\begin{figure*}[ht]
  \centering
  \includegraphics[width=0.98\linewidth]{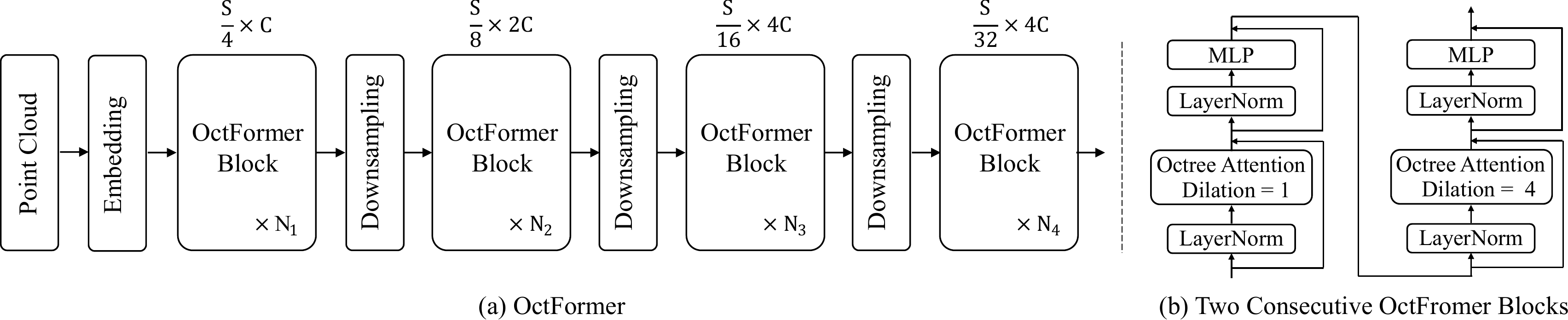}
  \caption{Overview.
  (a): The architecture of OctFormer. OctFormer consists of an Embedding module, a sequence of OctFormer blocks and downsampling modules.
  $S$ and $C$ denote the spatial resolution and channel of features, $N_i$ denotes the number of the corresponding OctFormer blocks.  
  (b): Two consecutive OctFormer blocks. 
  Each OctFormer block consists of an octree attention, an MLP, and two Layer Normalizations (LayerNorm). Two consecutive OctFormer blocks use dilations of 1 and 4 for their respective octree attentions. }
  \label{fig:octformer}
\end{figure*}

\paragraph{Point-based Networks}
Instead of rasterizing 3D shapes into regular voxels, point-based networks directly take raw point clouds as input.
Since point clouds are unordered and unstructured, these networks employ permutation-invariant operations~\cite{Qi2017a,Qi2017,Li2018}, continuous convolution kernals~\cite{Atzmon2018,Thomas2019,Fey2018}, or adaptive weights~\cite{Simonovsky2017,Wu2019a} to aggregate and update point features.
To query neighboring points, point-based networks construct a k-nearest-neighbor graph from input point clouds~\cite{Qi2017,Li2018,Xu2018,Simonovsky2017,Fey2018}.
In order to extract hierarchical features, point-based networks often rely on farthest point sampling~\cite{Qi2017} or grid-based sampling~\cite{Thomas2019,Hu2019} to downsample point clouds progressively.
Our OctFormer also takes point cloud as input, but our novel octree attention totally avoids expensive k-nearest-neighbour search and farthest point sampling.
As a result, OctFormer is much more efficient and outperforms previous point-based networks.

\paragraph{Vision Transformers}
Inspired by the great success of transformers in natural language processing~\cite{Vaswani2017}, ViT~\cite{Dosovitskiy2021} uses transformer-based networks for visual recognition.
ViT partitions the input image into non-overlapping regular patches, considers each patch as a token, and applies pure attentions to these tokens.
To extend ViT for dense prediction tasks such as image segmentation and detection, PVT~\cite{Wang2021c} and Swin Transformer~\cite{Liu2021a} introduce hierarchical network architectures from CNNs to vision transformers.
PVT proposes applying attention modules on downsampled feature maps to improve the efficiency of ViT when dealing with large images.
On the other hand, Swin Transformer introduces shifted-window attentions to restrict attentions in non-overlapping local windows.
Many follow-up works~\cite{Chu2021a,Dong2022,Wang2022a,Yang2021} further improve the attention designs with a similar network architecture.
OctFormer also has a hierarchical network architecture.
Our key innovation is a novel octree attention mechanism for point clouds, which utilizes variant local window shapes while maintaining a fixed number of points in each window for efficiency.

\paragraph{Point Cloud Transformers}
Following vision transformers, it is natural to explore the extension of transformers for point cloud understanding.
Point Cloud Transformer (PCT)~\cite{Guo2021} applies offset attentions to all point features for point cloud classification and segmentation.
Point-BERT~\cite{Yu2022} and Point-MAE~\cite{Pang2022} utilize standard transformers trained on point clouds for unsupervised pre-training.
3DETR~\cite{Misra2021} proposes an end-to-end scheme for point cloud detection with standard transformers.
However, these methods are limited to point clouds containing only a few thousand points due to the high computation and memory costs incurred by global attentions.

Point Transformer (PT)~\cite{Zhao2021} applies vector attentions to a local neighborhood of each point.
Although PT has lower memory costs than PCT, its computation cost is still high due to the usage of expensive farthest point sampling when doing pooling operations to downsample feature maps.
Point Transformer V2 (PTv2)~\cite{Wu2022} enhances PT's efficiency by substituting farthest point sampling with grid-based sampling and improves its performance further by utilizing grouped vector attentions and additional positional encoding multipliers.
The attention modules of both PT and PTv2 are applied independently to local neighborhoods of each point in a sliding window fashion.
Since there is no computation sharing among overlapping neighborhoods, significant computation resources are wasted.  \looseness=-1

To scale up point cloud transformers, SST~\cite{Fan2022}, SWFormer~\cite{Sun2022}, and Stratified Transformer~\cite{Lai2022} extend Swin Transformer~\cite{Liu2021a} for image understanding to point clouds by restricting attention modules to non-overlapping windows of point clouds.
Since the number of points in each local window differs greatly, SST and SWFormer group local windows with similar numbers of points together and process them in batch mode;
while Stratified Transformer leverages sophisticated GPU programs to combat this issue.
Similar to the scaling strategy of PVT~\cite{Wang2021c}, PatchFormer~\cite{Cheng2022} applies attention modules to patch features instead of point features;
Fast Point Transformer~\cite{Park2022} downsamples the point cloud into low-resolution voxels and restricts the attention modules to those voxels.
However, the performance of PatchFormer and Fast Point Transformer is worse than other contemporary point cloud transformers.
Unlike previous point cloud transformers, our OctFormer divides the input point clouds into groups containing an equal number of points, making it easy to parallelize and scale up.
Our OctFormer also achieves state-of-the-art performance on semantic segmentation and object detection on large-scale benchmarks.


\section{Octree-Based Transformers} \label{sec:method}

\begin{figure*}[ht]
  \centering
  \includegraphics[width=0.98\linewidth]{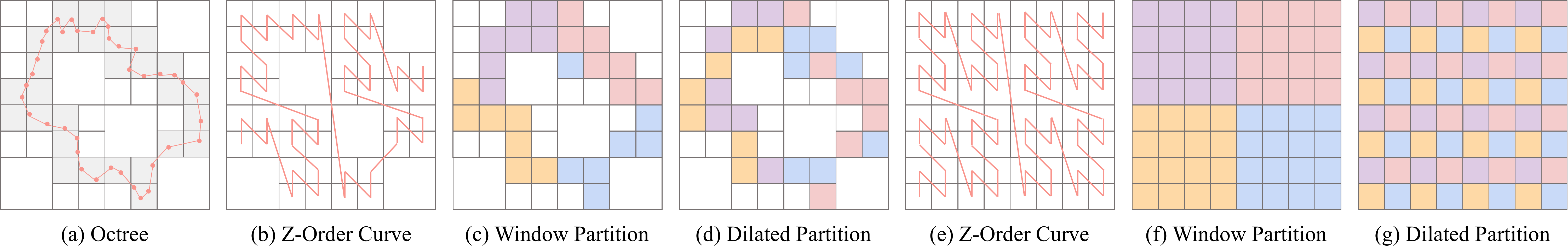}
  \caption{Window partition for the octree attention.
  Here 2D images are shown for a better illustration. 
  (a): An input point cloud sampled from a shape in red and the corresponding octree (quadtree). 
  Non-empty octree nodes are highlighted in gray.
  (b): Z-order curve at depth 3 of the octree.
  (c): A window partition generated by tensor reshaping and transposing corresponding to (b), with a point number of 7.
  The features are stored in a tensor following the order of non-empty octree nodes on the z-order curve.
  (d): A dilated partition with a point number of 7 and a dilation of 2.
  (e): Z-order curve covering the whole space.
  (f): A window partition corresponding to (e) with a point number of 16.
  (g): A dilated partition corresponding to (e) with a point number of 16 and a dilation of 4.
  }
  \label{fig:attention}
\end{figure*}

\paragraph{Overview}
The overview of OctFormer is shown in Figure~\ref{fig:octformer}.
Given an input point cloud, we first normalize it with a specified scale factor and convert it to an octree.
The initial features include average point positions, colors, and point normals (if provided) stored in non-empty octree leaf nodes.
Then an embedding module is used to downsample and project the initial features into a high-dimensional space.
Next, a sequence of OctFormer blocks and downsampling modules are alternately applied to generate high-level hierarchical features, which can be consumed by a lightweight Feature Pyramid Network (FPN)~\cite{Lin2017,Kirillov2019} for semantic segmentation and object detection.
The core of OctFormer is a novel octree attention module as shown in Figure~\ref{fig:octformer}-(b), which is elaborated in Section~\ref{subsec:attention}.
Other network components, including the embedding and downsampling module, the OctFormer block, and the detailed network configurations, are introduced in Section~\ref{subsec:network}.

\subsection{Octree Attention} \label{subsec:attention}

\paragraph{Attention}
Our octree attention is built upon the scaled dot-product attention proposed by~\cite{Vaswani2017}, which is widely used in transformers in NLP and vision.
We first review the attention module and introduce the observations motivating our octree attention.
Denote an input feature map as $X \in \mathbb{R}^{N \times C}$, where $N$ is the spatial number, and $C$ is the channel number.
Three learnable weight matrices $W_q $, $W_k $, and $W_v$ in $\mathbb{R}^{C \times D}$ are used to map $X$ to queries, keys and values.
Then the attention can be defined as follows:
\begin{equation}
  Attention(X) = softmax \Bigl( (X W_q) (X W_k)^T / \sqrt{D} \Bigr) (X W_v) .
  \label{eq:attn}
\end{equation}
Intuitively, the output can be regarded as a weighted sum of the values with weights dynamically computed from queries and the corresponding keys.
The multi-head attention can be implemented by concatenating $H$ independent attentions and merging the results with a linear layer, where $H$ denotes the head number.
In OctFormer, we adopt the multi-head attention by default, and we
directly use the term attention to refer to the multi-head attention for simplicity.

According to Equation~\ref{eq:attn}, the computation complexity of the attention is $\mathcal{O}(N^2)$, which is generally unaffordable when $N$ exceeds several thousand, whereas point clouds from large-scale scanned datasets, such as ScanNet~\cite{Dai2017a}, often contain over $100k$ points.
To scale up the attention, we leverage the strategy of window attentions~\cite{Liu2021a} to constrain the computation within non-overlapping local windows.
Denote the point number in each local window as $K$, then the computation complexity of window attentions is reduced to $\mathcal{O}(K^2 \cdot \frac{N}{K})$, which is linear to $N$.
A naive implementation is to follow vision transformers~\cite{Liu2021a} to partition point clouds into non-overlapping windows with 3D cubes~\cite{Fan2022,Sun2022,Lai2022}.
However, different from images, the point number in each window varies greatly.
Concretely, with a window size of 7 on ScanNet, the average point number is only 48, whereas the maximum point number is 343.
This issue severely hampers the efficient execution of attentions on GPUs.

Observing that the attention aggregates features via weighted averages with weights normalized by a \emph{softmax} function, we hypothesize that the attention is robust to the change of underlying window shapes.
And we verify this observation empirically via a pilot study.
Specifically, we use a ViT~\cite{Dosovitskiy2021} pretrained on a $16\times16$ partition of images defined within a square window and alter the window shape by randomly masking out 20\% of image patches, resulting in highly irregular windows.  
We utilize the codebase and weights provided by \emph{timm}~\cite{Wightman2019}.
After the masking operation, the accuracy of ViT on the validation set of ImageNet~\cite{Deng2009} drops only slightly, from 85.1\% to 84.2\%.
This observation motivates us to keep the point number in each local window constant while allowing the shape of local windows to vary freely, as opposed to simply using cubic windows.
By keeping the point numbers fixed in every window, the parallelized computation of GPU can be fully exploited, and the implementation can be greatly simplified.

\paragraph{Octree}
We adopt octrees to generate the required window partition and facilitate the hierarchical network architecture.
For each point cloud, we employ the parallel algorithm proposed by~\cite{Zhou2011} to build an octree on GPU.
After building the octree, octree nodes of the same depth are sorted with the shuffled keys~\cite{Wilhelms1992}.
Denote the integer coordinates of an octree node as $(x, y, z)$, and the $i^{th}$ bit of each coordinate as $x_i$, $y_i$, and $z_i$, the shuffled key in binary expression is defined as follows:
\begin{equation}
  Key(x, y, z)=x_1 y_1 z_1 x_2 y_2 z_2 \dots x_d y_d z_d ,
\end{equation}
where $d$ is the depth of the octree.
The value of a shuffled key represents the position on the 3D z-order curve.
We observe that octree nodes belong to one parent node, and more generally, octree nodes belonging to one subtree are contiguously stored in memory according to the z-order.
This property of octrees is the key to our efficient window partition.
A 2D illustration is shown in Figure~\ref{fig:attention}.
For an octree built from a point cloud marked in red color in Figure~\ref{fig:attention}-(a), the corresponding z-order curve in depth 3 of the octree is shown in Figure~\ref{fig:attention}-(b).
Since the octree nodes are sparse, the full z-order curve is shown in Figure~\ref{fig:attention}-(e) for reference.

\paragraph{Octree Attention}
With sorted shuffled keys of octrees, we can easily generate window partitions through tensor reshaping and transposing.
Note that only non-empty octree nodes contain feature vectors.
We can trivially filter out empty nodes with the information provided by octrees while preserving the original order.
An example is shown in Figure~\ref{fig:attention}-(a), where non-empty octree nodes are marked in gray color.
We stack all features in non-empty octree nodes into a 2D tensor following the order of sorted shuffled keys.
Denote the resulting feature tensor as $X$ with a shape of $(N, C)$, we first pad zeros to create a new tensor $\hat{X}$ with a shape of $(\hat{N}, C)$, so that $\hat{N}$ is divisible by the specified point number $K$, which is typically set to 32 in our experiments.
Then we generate the window partition by reshaping $\hat{X}$ to $(B, K, C)$, where $B$ equals $\hat{N} / K$ and denotes the total number of windows.
With this partition, we can implement our octree attention by applying the attention in Equation~\ref{eq:attn} to these $B$ windows in parallel while masking out the padded elements.
An example is shown in Figure~\ref{fig:attention}-(c), where $K$ and $N$ are equal to 7 and 28, respectively.

Although window attentions speed up the computation, they come with a reduced receptive field and a lack of information propagation among different windows.
To alleviate these issues, we further propose a dilated octree attention.
Denote the dilation as $D$, which is set to 1 or 4 in our experiments.
For the octree attention described above, its dilation can be regarded as 1.
When $D$ is larger than 1, we pad $X$ to $\tilde{X}$ so that the spatial number $\tilde{N}$ of $\tilde{X}$ is divisible by $K \times D$.
Next, we reshape $\tilde{X}$ to a tensor with shape $(\tilde{B}, K, D, C)$ where $\tilde{B}$ equals $ \tilde{N} / (K \times D)$.
Then we transpose it to a tensor with shape $(\tilde{B}, D, K, C)$, and flatten the first two dimensions to get a tensor with shape $(B, K, C)$, with which the attention in Equation~\ref{eq:attn} can also be directly applied.
An example is shown in Figure~\ref{fig:attention}-(c), where $K$ and $D$ are equal to 7 and 2, respectively.

Albeit our octree attention is designed for sparse voxels, it degenerates to the standard window attention~\cite{Liu2021a} and dilated attention~\cite{Hassani2022,Liu2022} in vision transformers when applied to full voxels or images with specific window size and dilation settings.
An example is shown in Figure~\ref{fig:attention}-(f)\&(g), where the point number is 16, and the dilation is 4.

\paragraph{Positional Encoding}
Positional encoding is essential for attentions to differentiate features defined at different positions.
A widely-used strategy is to add relative positional bias~\cite{Raffel2020,Liu2021a} to the attention defined in Equation~\ref{eq:attn}.
Denote the maximum window size as $W$; the relative positions of two arbitrary points within the same window lie in $[-W+1, W-1]$.
Naively extending relative positional bias to 3D requires $H \times (2W-1)^3$ trainable parameters, where $H$ is the head number of attention.
With our octree attention, the maximum window size is at least 50 when the dilation is larger than 4, which greatly increases the parameters of our OctFormer.
Therefore, we adopt the conditional positional encoding (CPE)~\cite{Chu2021,Dong2022,Wang2022a}, which dynamically generates the positional encoding conditioned on current features with a depthwise convolution.
Specifically, we apply the octree-based depthwise convolution provided by O-CNN~\cite{Wang2017} and the Batch Normalization~\cite{Loffe2015} to the input tensor $X$ as positional encoding before each attention module:
\begin{equation}
  X = X + batch\_norm(depth\_wise\_conv(X))
\end{equation}
With CPE, the network performance improves significantly while using fewer parameters than relative positional bias, as verified in our ablation study in Section~\ref{subsec:ablation}.

\begin{algorithm}[t]
  \caption{\small Pseudocode of Octree Attention in a PyTorch-like style.}
  \label{alg:code}
  \definecolor{codeblue}{rgb}{0.25,0.6,0.6}
  \definecolor{codekw}{rgb}{0.0, 0.0, 0.0}
  \lstset{
    backgroundcolor=\color{white},
    basicstyle=\fontsize{6.95pt}{6.95pt}\ttfamily\selectfont,
    columns=fullflexible,
    numbers=none,
    breaklines=true,
    captionpos=b,
    commentstyle=\fontsize{6.95pt}{6.95pt}\color{codeblue},
    keywordstyle=\fontsize{6.95pt}{6.95pt}\color{codekw},
  }
  \begin{lstlisting}[language=python]
  # x: input tensor with a shape of (N, C)
  # D: dilation for attention, set to 1 or 4 by default
  # P: point number in each window, set to 32 by default
  # attntion: an object of torch.nn.MultiheadAttention

  # apply conditional positional encoding
  x = x + batch_norm(depth_wise_conv(x))

  # window partition
  N, C = x.shape
  Nz = (P * D) - N % (P * D) # number of zeros for padding
  x = torch.cat([x, x.new_zeros(Nz, C)]) # pad zeros
  x = x.reshape(-1, P, D, C).transpose(1, 2).flatten(0, 1)

  # attention mask
  m = torch.cat([x.new_zeros(N), x.new_ones(Nz)]).bool()
  m = m.reshape(-1, P, D).transpose(1, 2).flatten(0, 1)

  # apply attention
  x = attntion(query=x, key=x, value=x, key_padding_mask=m)

  # reverse window partition
  x = x.reshape(-1, D, P, C).transpose(1, 2).reshape(-1, C)
  output = x[:N] # remove the padded elements
  \end{lstlisting}
\end{algorithm}

\paragraph{Summary}
Our octree attention is extremely easy to implement and much more efficient than previous point cloud attentions.
With open-sourced libraries on the web, we can implement our octree attention in 10 lines of code when the batch size is 1, as summarized in Algorithm~\ref{alg:code}.
The core of our octree attention can be implemented with the multi-head attention module provided by PyTorch -- \emph{torch.nn.MultiheadAttention}~\cite{Paszke2019}, which is highly optimized based on general matrix multiplication routines on GPU.
And tensor reshaping and transposing operators for window partition are also standard operators supported by PyTorch.
The number of zeros for padding in our octree attention is $(\tilde{N} - N)$, which is less than $K \times D$ (128 in our settings); thus the computation cost of padding is negligible.
In contrast, the implementation of other point cloud attentions~\cite{Sun2022,Lai2022,Fan2022} requires complex engineering and customized GPU programming.
We demonstrate the efficiency of our octree attention in Section~\ref{subsec:ablation}.

\subsection{Network Details}  \label{subsec:network}
In this section, we introduce the remaining details of OctFormer, including the embedding module, the OctFormer block, and the downsampling module, as shown in Figure~\ref{fig:octformer}.
Denote the initial spatial resolution of octree as $S$, the embedding module maps the input signal into a high-dimensional feature space and downsamples the spatial resolution by a factor of 4.
Following the embedding module are four network stages, and each network stage is composed of $N_i$ OctFormer blocks and a downsampling module, where $i$ denotes the stage index.
OctFormer blocks are used to process features; each downsampling module reduces the spatial resolution of features by a factor of 2 and increases the feature channel by a factor of 2 in each stage, except for the last stage, where the channel is kept as $4C$ to reduce the total parameter number of the network.

\paragraph{Embedding}
Instead of using a single convolution layer with a large kernel size and stride to instantiate the embedding module, as is done in ViT~\cite{Dosovitskiy2021}, we opt for several convolution layers with small kernel sizes, which has been shown to be able to stabilize the training process of transformers~\cite{Xiao2021}.
In OctFormer, we use a series of 5 octree-based convolution modules for the embedding module.
Each module consists of an octree convolution~\cite{Wang2017}, a Batch Normalization layer~\cite{Loffe2015}, and a ReLU activation function.
The kernel sizes and strides of these octree convolutions are $\{3, 2, 3, 2, 3\}$ and $\{1, 2, 1, 2, 1\}$, respectively.
The convolutions with a stride of 2 downsample the spatial resolution of tensors by a factor of 2.

\paragraph{OctFormer Block}
Following the standard transformer block design~\cite{Vaswani2017}, an OctFormer block consists of an octree attention introduced in Section~\ref{subsec:attention}, a multilayer perceptron (MLP), and residual connections, as shown in Figure~\ref{fig:octformer}-(b).
The MLP has two fully connected layers with a GELU activation function in between, and the expansion ratio of the hidden channel of MLP is set to 4.
Layer Normalization (LayerNorm)~\cite{Ba2016} is employed prior to each attention module and each MLP to stabilize the training.
The dilations of octree attention in two consecutive OctFormer blocks are set to 1 and 4, respectively.

\paragraph{Downsampling}
The downsampling module is implemented as an octree convolution with a kernel size of 2 and a stride of 2, followed by a Batch Normalization layer, which reduces the spatial resolution and increases the channel of a feature map by a factor of 2.

\paragraph{Network Settings}
By default, $C$ is set to 96, the block numbers are set to $\{2, 2, 18, 2\}$,  the head numbers of octree attentions are set to $1/16$ of the channel numbers, and the point numbers for window partition are set to 32.
The resulting OctFormer has a similar amount of trainable parameters (39M) as MinkowskiNet (38M)~\cite{Choy2019}.
The output of OctFormer is hierarchical features with four spatial resolutions, i.e., $\{ S/4, S/8, S/16, S/32 \}$, which can be conveniently integrated with a feature pyramid network (FPN)~\cite{Lin2017} for segmentation and detection.
And the last feature map can be averaged as the global feature for shape classification.

\section{Experiments} \label{sec:result}
In this section, we validate the effectiveness and efficiency of our OctFormer on 3D semantic segmentation and 3D object detection tasks.
We also discuss our design choices in the ablation study.
The experiments were conducted using 4 Nvidia 3090 GPUs with 24GB of memory.

\subsection{3D Semantic Segmentation} \label{sec:segmentation}
We first verify the efficacy of our OctFormer on 3D semantic segmentation.
The goal of this task is to predict the semantic label for each point in an input point cloud.

\paragraph{Dataset}
The experiments are conducted on the ScanNet~\cite{Dai2017a} dataset and the recently-proposed ScanNet200~\cite{Rozen2022} dataset.
ScanNet contains 1513 large-scale 3D scans of indoor scenes and includes 20 semantic categories.
The average point number of scans in ScanNet is $148k$.
ScanNet200 shares the same data as ScanNet, but has 200 semantic categories, making it more challenging.
We follow the standard data splits~\cite{Dai2017a} for training and evaluation, using 1201 scans for training, 312 scans for validation, and 100 scans for testing.
The testing labels are not publicly available, and testing results are obtained by submitting predictions to the official ScanNet website.
We use the mean intersection over union (mIoU) over all categories as the evaluation metric.

\begin{table}[t]
\centering
\tablestyle{8pt}{1.1}
\caption{Semantic Segmentation on ScanNet. 
\emph{Val.} and \emph{Test} denote the mIoU on the validation and testing set, respectively. The best results are marked in bold.
Our OctFormer achievs the best performance on the validation set, surpassing all point cloud transformers, sparse-voxel-based CNNs, and point-based networks. The mIoU of OctFormer without voting is shown in the parentheses for reference.
}
\begin{tabular}{l|cc}
  \toprule
  Method                                      & Val.          & Test       \\
  \midrule
  3DMV~\cite{Dai2018a}                        & -             & 48.4       \\
  PanopticFusion~\cite{Narita2019}            & -             & 52.9       \\
  PointNet++~\cite{Qi2017}                    & 53.5          & 55.7       \\
  SegGCN~\cite{Lei2020a}                      & -             & 58.9       \\
  JointPoint~\cite{Chiang2019}                & 69.2          & 63.4       \\
  RandLA-Net~\cite{Hu2019}                    & -             & 64.5       \\
  PointConv~\cite{Wu2019a}                    & 61.0          & 66.6       \\
  PointASNL~\cite{Yan2020a}                   & 63.5          & 66.6       \\
  KPConv~\cite{Thomas2019}                    & 69.2          & 68.6       \\
  FusionNet~\cite{Zhang2020}                  & -             & 68.8       \\
  JSENet~\cite{Hu2020}                        & -             & 69.9       \\
  SparseConvNet~\cite{Graham2018}             & 69.3          & 72.5       \\
  MinkowskiNet~\cite{Choy2019}                & 72.2          & 73.6       \\
  LargeKernel~\cite{Chen2022}                 & 73.2          & 73.9       \\
  O-CNN~\cite{Wang2017}                       & 74.5          & 76.2       \\
  Mix3D~\cite{Nekrasov2021}                   & 73.6          & \tb{78.2}  \\
  \midrule
  Point Transformer~\cite{Zhao2021}           & 70.6          & -          \\
  Fast Point Transformer~\cite{Zhao2021}      & 72.1          & -          \\
  Stratified Transformer~\cite{Lai2022}       & 74.3          & 73.7       \\
  Point Transformer V2~\cite{Wu2022}          & 75.4          & 75.2       \\
  OctFormer (ours)                            & \tb{75.7} (74.5)     & 76.6  \\
  \bottomrule
\end{tabular}
\label{tab:scannetv2}
\end{table}

\paragraph{Settings}
OctFormer is used as the backbone to extract hierarchical features, and a lightweight FPN~\cite{Lin2017} is used as the segmentation head.
The FPN first projects multiscale features with a convolution layer with a kernel size of 1 to make the channels of features equal to 168, then upsamples the features with the nearest interpolation by a factor of 2 and merges consecutive features by addition.
The final output feature is processed by a convolution with a kernel size of 3 and upsampled to each point with the nearest neighbor interpolation, with which
an MLP with one hidden layer is used to predict the point categories.
For both ScanNet and ScanNet200, the training settings are the same, except that the output channel is 20 for ScanNet and 200 for ScanNet200.
We employ an AdamW optimizer~\cite{Loshchilov2017} to train the network for 600 epochs with a batch size of 16 and a weight decay of 0.05.
The initial learning rate is set as 0.006 and decreases by a factor of 10 after 360 and 480 epochs, respectively.
The input point clouds are first normalized with a scale factor of 0.01m and then encoded by octrees with a depth of 11.
The initial features include colors, normals, and point coordinates.
The data augmentations include random rotation in $[-180^\circ, 180^\circ]$, random scaling in $[0.75, 1.25]$, random translation in $[-0.1, 0.1]$, random elastic deformations following~\cite{Choy2019}, and random spherical cropping following~\cite{Lai2022}.

\begin{table}[t]
\centering
\tablestyle{6pt}{1.1}
\caption{Semantic Segmentation on ScanNet200. \emph{Head}, \emph{Common} and \emph{Tail} denote mIoUs on three smaller groups containing 66, 68 and 66 categories of ScanNet200, \emph{All} denotes mIoU on all 200 categories. The best results are marked in bold. Our OctFormer trained from scratch acheives the best performance on all groups and are significantly better than previous state-of-the-art methods, even those with pretraining.}
\begin{tabular}{l|ccc|c}
  \toprule
  Method                           & Head       & Common     & Tail        & All       \\
  \midrule
  CSC-Pretrain~\cite{Hou2021}      & 45.5       & 17.1       & 7.9         & 24.9      \\
  MinkowskiNet~\cite{Choy2019}     & 46.3       & 15.4       & 10.2        & 25.3      \\
  SupCon~\cite{Khosla2020}         & 48.6       & 19.2       & 10.3        & 26.0      \\
  LGround~\cite{Rozen2022}         & 48.5       & 18.4       & 10.6        & 27.2      \\
  OctFormer (ours)                 & \bf{53.9}	& \bf{26.5}  & \bf{13.1}   & \bf{32.6} \\
  \bottomrule
\end{tabular}
\label{tab:scannet200}
\end{table}

\begin{figure}[t]
  \centering
  \includegraphics[width=\linewidth]{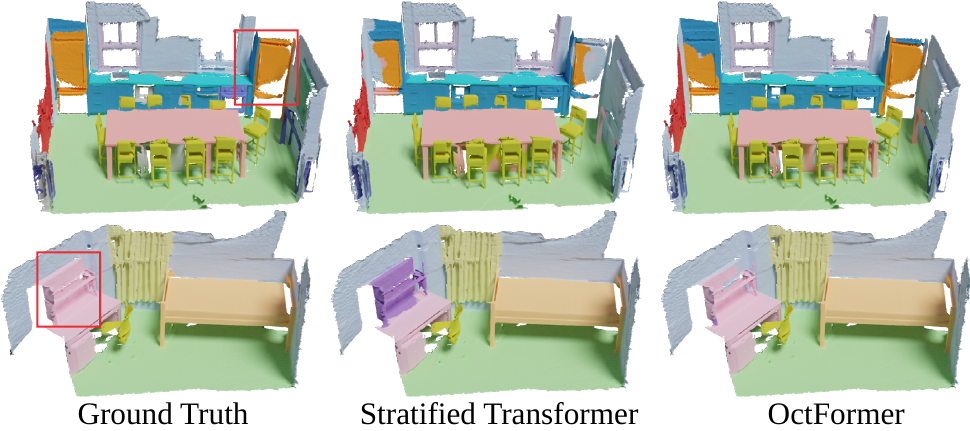}
  \caption{Visual comparison on ScanNet. The results of OctFormer are more faithful to the ground truth, as highlighted in the rectangle regions.}
  \label{fig:scannet}
\end{figure}

\paragraph{Results on ScanNet}
We compare our OctFormer with a series of previous state-of-the-art methods on ScanNet and summarize the validation and testing mIoUs in Table~\ref{tab:scannetv2}.
The training of our OctFormer takes 15 hours and consumes 13GB of memory using 4 Nvidia 3090 GPUs.
Our OctFormer achieves a mIoU of \emph{74.5} on the validation set without voting.
Since Stratified Transformer~\cite{Lai2022}  employs block-wise prediction and Point Transformer V2~\cite{Wu2022} employs the voting strategy to improve the performance, we also adopt the voting strategy, which results in a mIoU of 75.7 as shown in Table~\ref{tab:scannetv2}.
Clearly, our OctFormer achieves the best performance on the validation set among all methods.
Specifically, OctFormer outperforms previous point cloud transformers with a validation mIoU higher than Stratified Transformer~\cite{Lai2022} by 1.4 and Point Transformer~\cite{Zhao2021} by 5.1.
OctFormer also surpasses sparse-voxel-based CNNs with a validation mIoU higher than SparseConvNet~\cite{Graham2018} by 6.4, MinkowskiNet~\cite{Choy2019} by 3.5, and LargeKernel~\cite{Chen2022} by 2.5.
Additionally, OctFormer significantly outperforms point-based networks, including PointNet++~\cite{Qi2017} and KPConv~\cite{Thomas2019}.
On the testing set, our OctFormer achieves the second-best mIoU, as Mix3D currently ranks the first.
However, our OctFormer achieves better mIoU than Mix3D on the validation set.
One possible explanation is that Mix3D employs an additional mixup data augmentation and post-processing to further improve the testing results, as mentioned in~\cite{Nekrasov2021}.
A visual comparison with Stratified Transformer is shown in Figure~\ref{fig:scannet}, which demonstrates that OctFormer can produce more faithful results in detailed regions.
 \looseness=-1


\paragraph{Results on ScanNet200}
Since object categories in ScanNet200 are highly imbalanced, the 200 categories of ScanNet200 are further split into three smaller groups of 66, 68 and 66 categories~\cite{Rozen2022} according to the label frequency in the training set, which are denoted as Head, Common and Tail respectively.
The mIoUs of these small groups and all categories on the testing set are reported in Table~\ref{tab:scannet200}.
Among the listed methods in Table~\ref{tab:scannet200}, MinkowskiNet and our Octformer are trained from scratch with random initialization, while CSC-Pretrain~\cite{Hou2021} and SupCon~\cite{Khosla2020} use additional data to pretrain the network with contrastive learning, and LGround~\cite{Rozen2022} is a newly-proposed language-driven pre-training method based on a pretrained large vision-language CLIP model~\cite{Radford2021}.
As shown in Table~\ref{tab:scannet200}, our Octformer trained from scratch without additional data is significantly better than these competitors and also consistently better in the three small groups, which verifies the effectiveness of our Octformer.
Specifically, the mIoU of our OctFormer is higher than MinkowskiNet by 7.3 and higher than LGround by 5.4, although LGround is pretrained with the help of a powerful CLIP model.

\begin{figure}[t]
  \centering
  \includegraphics[width=0.8\linewidth]{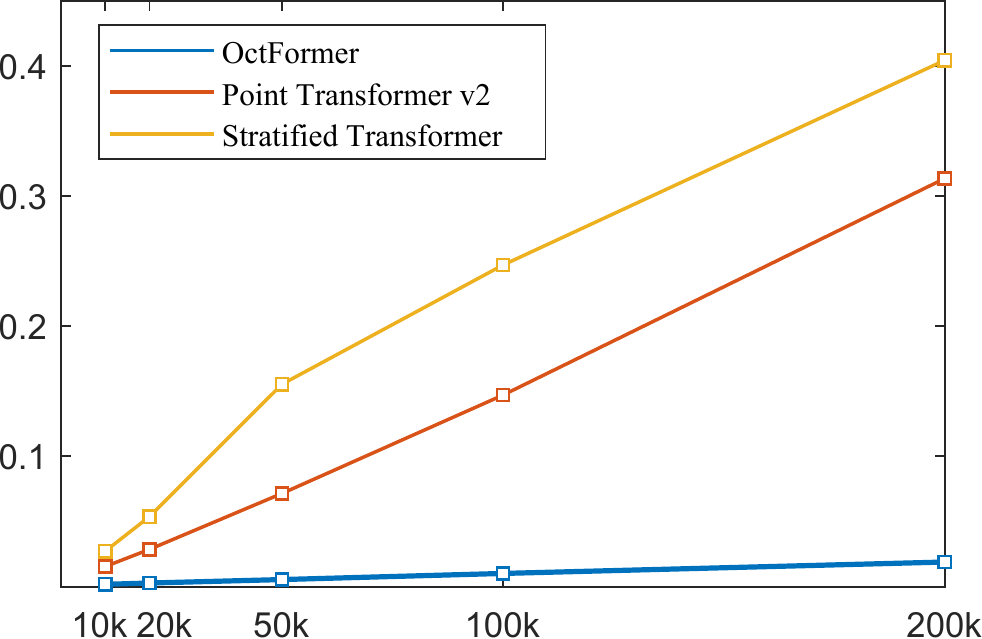}
  \caption{Efficiency comparisons. The horizontal axis represents the spatial number of an input tensor, and the vertical axis represents the running time in seconds.
  Our OctFormer runs significantly faster and is over 17 times faster than Point Transformer v2 and Stratified Transformer when the spatial number is $200k$.   }
  \label{fig:efficiency}
\end{figure}

\subsection{Ablation Studies and Discussions} \label{subsec:ablation}

In this section, we justify key design choices of Octformer and compare the efficiency of OctFormer with other point cloud transformers on top of the semantic segmentation on ScanNet in Section~\ref{sec:segmentation}.

\begin{table*}[t]
  \centering
  \caption{Ablation studies on semantic segmentation on ScanNet.
  \emph{mIoU} and \emph{Loss} denote the validation mIoU and the training loss. The training loss is used as a reference of the expresiveness of networks, which is multiplied by 100 for better display.
  Best results are marked in bold. }
  \label{tab:ablations}
  \subfloat[
  Model size. A large model can steadily improve the performance.
  \label{tab:network}]{
  \centering
  \begin{minipage}{0.29\linewidth}{
    \begin{center}
    \tablestyle{8pt}{1.1}
    \begin{tabular}{cccc}
      Name    & Model Size  & Loss       & mIoU       \\
      \hline
      Small   & 18M         & 10.2       & 74.0      \\
      Base    & 39M         & 8.12       & 74.5     \\
      Large   & 156M        & \tb{7.22}  & \tb{74.6}  \\  
    \end{tabular}
    \end{center}}
  \end{minipage}}
  \hfill
  \subfloat[
  Positional encoding. CPE is effective and essential for OctFormer.
  \label{tab:pos_enc}]{
  \begin{minipage}{0.29\linewidth}{
    \begin{center}
    \tablestyle{14pt}{1.1}
    \begin{tabular}{lcc}
      Type         & Loss      & mIoU \\
      \hline
      w/o          & 21.7      & 66.5 \\
      cRPE         & 8.43        & 73.9 \\
      CPE          & \bf{8.12}   & \bf{74.5} \\
      \end{tabular}
    \end{center}}
  \end{minipage}}
  \hfill
  \subfloat[
  Input voxel size. OctFormer is efficient enough to take fine voxels for better results.
  \label{tab:voxel_size}]{
  \begin{minipage}{0.295\linewidth}{
    \begin{center}
    \tablestyle{14pt}{1.05}
    \begin{tabular}{ccc}
      Voxel Size    & Loss        & mIoU      \\
      \hline
      4cm           & 12.6        & 70.7      \\
      2cm           & 9.08        & 73.6        \\
      1cm           & \bf{8.12}   & \bf{74.5} \\
    \end{tabular}
    \end{center}}
  \end{minipage}}
  \\
  \centering
  \vspace{1em}
  \subfloat[
    Point number in each window. Large point number increases network capacity.
    \label{tab:pt_num}]{
    \begin{minipage}{0.29\linewidth}{
    \begin{center}
    \tablestyle{8pt}{1.1}
    \begin{tabular}{cccc}
      Num.         & Win. Size & Loss        & mIoU       \\
      \hline
      16           &  7        &  8.24       & 73.8       \\
      32           &  10       &  8.12       & \bf{74.5}  \\
      48           &  12       &  8.10       & 74.3       \\
      64           &  15       &  \tb{7.92}  & 74.4       \\
      \end{tabular}
    \end{center}}
  \end{minipage}}
  \hfill
  \subfloat[
  Dilation. Large dilation also increases network capacity.
  \label{tab:dilation}]{
  \begin{minipage}{0.29\linewidth}{
    \begin{center}
    \tablestyle{12pt}{1.1}
    \begin{tabular}{ccc}
      Dilation   & Loss & mIoU \\
      \hline
      1           &  8.29       & 74.2 \\
      2           &  8.17       & 74.4 \\
      4           &  8.12       & \bf{74.5} \\
      8           &  \bf{8.01}  & 74.1 \\
      \end{tabular}
    \end{center}}
  \end{minipage}}
  \hfill
  \subfloat[
  Data augmentations. Strong data augmentations are helpful for a good mIoU.
  \label{tab:data_aug}]{
  \begin{minipage}{0.295\linewidth}{
    \begin{center}
    \tablestyle{12pt}{1.05}
    \begin{tabular}{lcc}
      Augmentation     & Loss         & mIoU \\
      \hline
      w/o              & \bf{0.58}    & 53.8 \\
      +Affine          & 5.59         & 72.7 \\
      +Crop            & 6.27         & 73.6 \\
      +Elastic         & 8.12         & \bf{74.5} \\
    \end{tabular}
    \end{center}}
  \end{minipage}}
\end{table*}

\paragraph{Efficiency}
Global point cloud transformers~\cite{Guo2021,Yu2022,Pang2022} can only process point clouds containing several thousand points,
thus we omit the comparisons with them and focus on the comparisons with efficient point cloud transformers proposed recently, including Stratified Transformer~\cite{Lai2022} and Point Transformer v2~\cite{Wu2022}.
Stratified Transformer extends window attentions~\cite{Liu2021a} to point clouds with cubic windows~\cite{Fan2022,Sun2022} and leverages stratified sampling to improve its performance.
Point Transformer v2 applies attentions to k nearest neighbors of each point in a sliding-window fashion.
Since the network configurations vary greatly,  we record the running time of one \emph{single} transformer block on an Nvidia 3090 GPU to eliminate the influence of uncontrolled factors,
We choose the input tensor's spatial number  from $\{10k, 20k, 50k, 100k, 200k\}$ and set the channel as 96.
For the attention modules, we set the head number to 6, and set the point number and neighborhood number to 32 for OctFormer and Point Transformer v2.
Since the point number is variant in each window for Stratified Transformer, we set the window size to 6 so that the average point number is about 32.

The results are shown in Figure~\ref{fig:efficiency}.
It can be seen that although the computation complexities of the three methods are all linear, our OctFormer runs significantly faster than  Point Transformer v2 and Stratified Transformer.
OctFormer runs over 17 times faster than the other two methods when the spatial number of the input tensor is $200k$.
The key reason for the efficiency of our Octformer is that our novel octree attention mainly leverages standard operators supported by deep learning frameworks, which is further based on general matrix multiplication routines on GPUs and has been optimized towards the computation limit of GPUs. 
Conversely, the point number in each window of Stratified Transformer is highly unbalanced, making it challenging for efficiency tuning even with hand-crafted GPU programming.
Although the neighborhood number of Point Transformer v2 is fixed, the sliding window execution pattern wastes considerable computation that could have been shared.

We also compare the efficiency of the whole network as shown in Figure~\ref{fig:teaser}.
We record the time of one forward pass of each network on a Nivida 3090 GPU, taking a batch of $250k$ points.
The speed of our Octformer-Small is slightly faster than MinkowskiNet, and faster than Point Transformer V2 by 3 times and Stratified Transformer by  20 times.
It is worth mentioning that our OctFormer takes point clouds quantized by a voxel size of 1cm as input, whereas the other networks takes point clouds quantized by a voxel size of 2cm.


\paragraph{Model Size}
We denote the default OctFormer as OctFormer-Base and design two variants with different amounts of trainable parameters, including OctFormer-Small with half of the parameters of OctFormer-Base and OctFormer-Large with 4 times of parameters of OctFormer-Base.
The detailed settings are listed as follows:
\begin{itemize}[leftmargin=16pt,itemsep=2pt]
  \item[-] OctFormer-Small: $C=96$, \, block numbers =  $\{2, 2, 6, 2\}$;
  \item[-] OctFormer-Large: $C=192$, block numbers = $\{2, 2, 18, 2\}$.
\end{itemize}
We test the performances of these models and summarize the results in Table~\ref{tab:network}.
It can be seen that as the model size increases, the mIoU on the validation set and the training loss improve steadily.
The training loss can be used as a metric that directly reflects the network's expressiveness.
These results indicate that our OctFormer scales up well with increasing model sizes.
Note that our OctFormer-Small with only 18M parameters achieves a mIoU of 74.1 (\emph{75.2 with voting}) on the validation set of ScanNet and already surpasses previous sparse-voxel-based CNNs, like Mix3D~\cite{Nekrasov2021} and MinkowskiNet~\cite{Choy2019} with 38M parameters, as shown in Table~\ref{tab:scannetv2}.

\paragraph{Positional Encoding}
Here we investigate the effect of conditional positional encoding (CPE)~\cite{Chu2021} and report the results in Table~\ref{tab:pos_enc}.
After removing CPE, denoted as \emph{w/o} in the second row of Table~\ref{tab:pos_enc}, the mIoU decreases by 8.0, and the training loss also increases greatly, which verifies that the positional encoding is crucial for OctFormer to perceive positional information.
Additionally, we also test the contextual relative positional encoding (cRPE) proposed by Stratified Transformer~\cite{Lai2022} with OctFormer and get a mIoU of 73.9, which is much better than the model without positional encoding but still slightly worse than CPE.
Although it is possible to further improve the mIoU by tuning the parameters of cRPE, like the quantization size of cRPE, cRPE is tightly coupled with the attention module, and its implementation is more complex than CPE.

\paragraph{Input Voxel Size}
Octrees are used to rasterize point clouds to sparse voxels, and it is known that finer voxels retain more information, which can help networks achieve better results.
We train OctFormer on ScanNet using three different voxel sizes and report the results in Table~\ref{tab:voxel_size}.
OctFormer achieve the best results with a voxel size of 1cm, improving the mIoU by 0.9 compared with a voxel size of 2cm.
MinkowskiNet~\cite{Choy2019} and Stratified Transformer~\cite{Lai2022} are typically trained on ScanNet with a voxel size of 2cm by default.
However, they run out of memory on Nvidia 3090 GPUs with 24GB memory when using a voxel size of 1cm, requiring further parameter tuning to produce reasonable results.
In contrast, our OctFormer, even OctFormer-Large, is efficient enough to be trained with a voxel size of 1cm within 24GB GPU memory.

\paragraph{Point Number in Each Window}
The point number $K$ in each window is closely related to the window size and the receptive field of octree attentions.
We train OctFormer with a point number chosen from $\{16, 32, 48, 64\}$ and report the results in Table~\ref{tab:pt_num}.
With a fixed point number, the window sizes are variable, and we calculated the average window sizes in Table~\ref{tab:pt_num}.
The window size in Stratified Transformer~\cite{Lai2022} is 5, and the average window size of OctFormer with a default point number of 32 is 10, which greatly increases the receptive field of the network.
It is evident from Table~\ref{tab:pt_num} that the training loss progressively decreases as the point number increases.
Although the mIoU of OctFormer with a point number of 64 is slightly worse than with a point number of 32, the training loss with a point number of 64 is lower, which is a sign of overfitting.
Therefore, we can conclude that the expressiveness or capacity of the network increases with the point number of octree attention.

\paragraph{Dilation}
Dilated octree attentions further increase the receptive field by using dilated local windows, which are controlled by the hyperparameter dilation $D$.
We train OctFormer with a dilation chosen from $\{1, 2, 4, 8\}$ and report the results in Table~\ref{tab:dilation}.
Similarly, we observe that the mIoU improves until the dilation reaches 4.
And the training loss decreases as the dilation increases.
When the dilation is 8, the training loss is the best, indicating that the capacity of the network increases with the dilation of octree attention.

\paragraph{Data Augmentation}
We inspect the influence of different data augmentations on the mIoU by progressively adding each of them and summarize the results in Table~\ref{tab:data_aug}.
Without data augmentations, we observe severe overfitting, leading to much lower mIoU.
As we add more data augmentations, the mIoU gradually increases.
Although Mix3D~\cite{Nekrasov2021} leverages an additional 3D mixup augmentation to achieve the best performance on the testing set of ScanNet, we chose to use only the data augmentations in Point Transformer V2~\cite{Wu2022} and Stratified Transformer~\cite{Lai2022} for a fair comparison with point cloud transformers.

\subsection{3D Object Detection}
In this section, we  validate the effectiveness of our OctFormer on 3D object detection.
The goal is to predict 3D bounding boxes and the corresponding categories of objects contained in an input point cloud. \looseness=-1

\paragraph{Dataset}
We perform 3D object detection on the SUN RGB-D dataset~\cite{Song2015}, which contains about 10k single-view RGB-D scans of indoor scenes.
The annotations of RGB-D scans consist of oriented 3D bounding boxes of 37 categories.
Following previous settings~\cite{Qi2019,Rukhovich2022}, we use 5285 and 5050 RGB-D scans for training and validation, respectively, and report the average precision (mAP) under IoU thresholds of 0.25 and 0.5, denoted as mAP@0.25 and mAP@0.5, for the ten most common categories for performance comparisons.

\paragraph{Settings}
We adapt FCAF3D~\cite{Rukhovich2022} with our OctFormer for object detection.
FCAF3D currently achieves top performance on the SUN RGB-D detection benchmark.
The backbone of FCAF3D is a ResNet with 34 sparse convolution layers built upon MinkowskiNet~\cite{Choy2019}.
The detection head of FCAF3D is an FPN~\cite{Lin2017}, which is widely used in object detection.
We replace the original backbone network of FCAF3D with our OctFormer while keeping other components fixed.
To be compatible with the configurations of FCAF3D, we further add two octree convolution modules with kernel sizes $\{3, 2\}$ and stride $\{1, 2\}$ before the input of OctFormer so that the resolutions of the resulting features are similar to the original backbone of FCAF3D.
We perform experiments with the MMDetection3D framework.
We employ an AdamW optimizer~\cite{Loshchilov2017} to train the network for 18 epochs with a batch size of 32 and a weight decay of 0.01.
The initial learning rate is set to 0.001 and decreases by a factor of 10 after 12 and 16 epochs, respectively.
The input point clouds are first normalized with a voxel size of 0.01m and then encoded by octrees with a depth of 12, while the initial input features are point colors with 3 channels.
The data augmentations include random rotation in $[-30^\circ, 30^\circ]$ along the upright axis, random scaling in $[0.85, 1.15]$, and random translation in $[-0.1, 0.1]$, which are similar to FCAF3D.

\begin{table}[t]
  \centering
  \tablestyle{8pt}{1.1}
  \caption{3D Object detection on SUN RGB-D. The mAP@0.25 and mAP@0.5 denote the mean average precision under IoU threshold of 0.25 and 0.5, respectively.
  We run the experiments 5 times and report the maximum and average metric values; the average values are shown in brackets.
  By replacing the backbone of FACF3D with our OctFormer, we achieve the best performance on mAP@0.5, surpassing all pervious state-of-the-art methods.}
  \begin{tabular}{l|cc}
    \toprule
    Method                      & mAP@0.25       & mAP@0.5       \\
    \midrule
    VoteNet~\cite{Qi2019}       & 57.7           & -             \\
    MLCVNet~\cite{Xie2020a}     & 59.8           & -             \\
    3DETR~\cite{Misra2021}      & 59.1           & 32.7          \\
    H3DNet~\cite{Zhang2020a}    & 60.1           & 39.0          \\
    BRNet~\cite{Cheng2021}      & 61.1           & 43.7          \\
    HGNet~\cite{Chen2020b}      & 61.6           & -             \\
    VENet~\cite{Xie2021a}       & 62.5           & 39.2          \\
    GroupFree~\cite{Liu2021e}   & 63.0 (62.6)    & 45.2 (44.4)   \\
    CAGroup3D~\cite{Wang2022b}  & \tb{66.8 (66.4)}    & 50.2 (49.5)   \\
    \midrule
    FCAF3D~\cite{Rukhovich2022} & 64.2 (63.8)         & 48.9 (48.2)      \\
    OctFormer (ours)      & 66.2 (65.7)    & \tb{50.6 (50.2)} \\
    \bottomrule
  \end{tabular}
  \label{tab:detection}
\end{table}

\paragraph{Results}
We do comparisons with previous state-of-the-art methods on SUN RGB-D and summarize the results in Table~\ref{tab:detection}.
We run our OctFormer 5 times and report the average and maximum performance to reduce the effect of random fluctuations.
Each training process takes approximately 3.5 hours on four Nvidia 3090 GPUs, and the maximum GPU memory consumption is less than 8GB.
It can be seen from Table~\ref{tab:detection} that our method achieves the best results in terms of mAP@0.5.
Compared with FCAF3D, the only difference is replacing the original backbone of FCAF3D with our OctFormer, and the metrics mAP@0.25 and mAP@0.5 directly increase by 1.9 and 2.0, respectively, which clearly demonstrates the effectiveness of our OctFormer over sparse-voxel-based CNNs.
Among the listed methods in Table~\ref{tab:detection}, 3DETR~\cite{Misra2021} and GroupFree~\cite{Liu2021e} are based on point cloud transformers.
Our OctFormer achieves significantly better performance than them.
Specifically, the mAP@0.25 of our OctFormer is higher than 3DETR and GroupFree by \emph{6.6} and \emph{3.1}, and mAP@0.5 is higher by \emph{17.5} and \emph{5.8}, respectively.
Lastly, it should be noted that the contributions of CAGroup3D~\cite{Wang2022b} are orthogonal to our OctFormer.
CAGroup3D mainly improves FCAF3D by designing the detection head, whereas our goal is to demonstrate the advantages of OctFormer over other backbones by replacing the backbone of FACF3D.
Nevertheless, the average mAP@0.5 of our method is still higher than CAGroup3D by 0.7.

\section{Conclusion} \label{sec:conclusion}

We propose OctFormer, a general and effective backbone for 3D point cloud understanding.
The core of OctFormer contains a novel octree attention and its dilated variant.
Our octree attention is extremely easy to implement and runs significantly faster than other point cloud attentions.
OctFormer demonstrates great efficiency and achieves state-of-the-art performance on several benchmarks, including semantic segmentation on ScanNet and ScanNet200 and 3D object detection on SUN RGB-D.

The field of 3D deep learning is rapidly evolving, and it is inevitable that the performance of OctFormer on the leaderboards will be surpassed by future works sooner or later.
However, our novel octree attentions, with their highly simplified implementation and super efficiency, along with our unified network design, will make 3D transformers more accessible to a broader audience and open up many exciting possibilities for the future, including pretraining of large-scale general 3D models, cross-modality training with images or languages, and more.

The limitations and future works are discussed as follows:

\paragraph{Small-Scale Dataset}
One limitation of OctFormer is that it is prone to overfit on small-scale datasets.
We test our OctFormer on the ModelNet40 classification, which only contains about 9k shapes for training and 2k for testing.
We use the average features produced by the last stage of OctFormer as the shape features for classification and get an accuracy of 92.7\% on the testing set without voting.
Although OctFormer is superior to the point transformer in Point-BERT~\cite{Yu2022} with an accuracy of 91.2\%, it is still worse than PointMLP~\cite{Ma2022} with an accuracy of 94.1\%.
Point-BERT~\cite{Yu2022} and Point-MAE~\cite{Pang2022} improve the accuracy of their point transformer to 93.8\% with unsupervised pretraining.
We believe similar unsupervised pretraining techniques can also help  OctFormer to combat the overfitting issue to achieve better performance on small-scale datasets, which is left as future work.  \looseness=-1

\paragraph{Positional Encoding}
As verified in our ablation studies, positional encoding is essential for transformers, and we currently adopt CPE~\cite{Chu2021} as the positional encoding of OctFormer.
Although CPE is effective, it hurts the flexibility of OctFormer due to the dependency on octree-based depth-wise convolutions.
In the future, we will explore the possibility of other positional encodings, e.g., MLP-based positional encoding that encodes the relative positional information with MLPs.

\paragraph{Cross Attention}
The proposed octree attention is essentially a self-attention, which is mainly used by OctFormer as a building block of an encoder network.
Additionally, the cross attention is also indispensable for learning complex relationships between queries and keys and has been successfully used in the decoder of 3DETR~\cite{Misra2021} for 3D object detection.
We regard the extension of octree attention to the cross-attention as another future work.

\paragraph{3D Generation}
We focus on using OctFormer for point cloud understanding tasks in this paper.
It is interesting to apply our OctFormer, or other network architectures designed with our octree attention, for 3D content creation with cross-modality training, e.g., 3D shape generation conditioned on images, sketches, or texts.

\begin{acks}
This work was supported in part by National Key R\&D Program of China 2022ZD0160801.
We also thank the anonymous reviewers for their valuable feedback.
\end{acks}

\bibliographystyle{ACM-Reference-Format}
\bibliography{src/ref/reference}


\end{document}